\documentclass[acmtog]{acmart} 

\AtBeginDocument{%
  }

\acmSubmissionID{1180}

\begin{document}

\title{Real-Time Human-Centric World Modeling for Upper-Body Human-Object Interaction}


\author{Chaonan Ji}
\affiliation{%
  \institution{Tongyi Lab}
  \city{Beijing}
  \country{China}}
\email{18801313326@163.com}

\author{Jinwei Qi}
\affiliation{%
  \institution{Tongyi Lab}
  \city{Beijing}
  \country{China}}
\email{jinwei.qjw@alibaba-inc.com}

\author{Peng Zhang}
\affiliation{%
  \institution{Tongyi Lab}
  \city{Beijing}
  \country{China}}
\email{futian.zp@alibaba-inc.com}

\author{Bang Zhang}
\affiliation{%
  \institution{Tongyi Lab}
  \city{Beijing}
  \country{China}}
\email{bangzhang@gmail.com}


\begin{abstract}
  We present a real-time human-centric world model for upper-body interactive generation, aiming to synthesize coherent local world dynamics centered on a person, where coordinated body, hand, and facial motions evolve jointly with controllable human-object discrete interaction. To this end, we adopt a continuous-discrete joint control scheme with two complementary components: a continuous human state and a discrete interaction state. For continuous human-state control, we introduce a unified implicit representation based on multi-scale motion encoding, in which motion latents from the upper body, hands, and face are fused into a shared latent space. This multi-scale design improves expressiveness across different spatial scales, captures fine-grained human dynamics more effectively, and enables direct control without explicit retargeting. For discrete object interaction-state control, we represent object contact using a small set of language-encoded discrete interaction states, where text serves as an explicit interaction-state command, such as \emph{no contact} or \emph{grasp}, rather than an open-ended generation prompt, and we further construct a dedicated rendering pipeline for human-object interaction data to supervise such discrete interaction states. By combining continuous implicit human-state control with discrete interaction-state control, our model enables precise modeling of how a person moves and interacts with the local environment, including controllable changes to nearby scene states. Finally, we distill the model for efficient streaming real-time inference, achieving 25 FPS on two H100 GPUs. Experiments demonstrate improved fine-grained motion fidelity, more realistic hand-object coordination, and effective real-time interaction, establishing a practical step beyond motion reproduction toward real-time human-centric world modeling.
\end{abstract}


\begin{teaserfigure}
  \includegraphics[width=\textwidth]{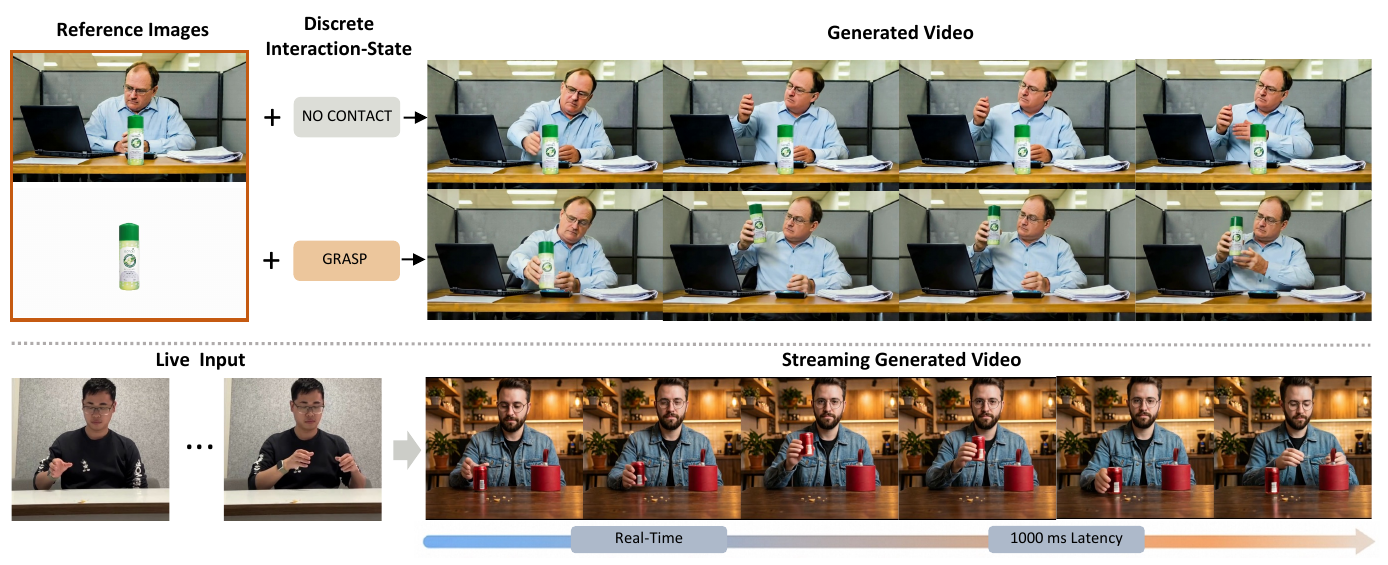}
  \caption{Visual results. Conditioned on a reference image, an object image, a driving input, and a discrete interaction-state command, our method synthesizes controllable human--object interaction results. The driving input is omitted in the upper examples for brevity and shown only in the streaming real-time results below, where our system supports streaming generation from live driving input at 25 FPS with about 1000 ms latency.}
  \label{fig:teaser_v2}
\end{teaserfigure}


\maketitle
\section{Introduction}
Recent advances in human video generation \cite{DBLP:conf/nips/SiarohinLT0S19,DBLP:journals/corr/abs-2407-03168,DBLP:conf/iclr/ZhaoXSXZ0LSL25,DBLP:conf/siggrapha/MaLWPHYZCS0C24,DBLP:conf/siggrapha/SongXZXGLZL25,cheng2025wananimateunifiedcharacteranimation} have substantially improved the quality and realism of upper-body human video synthesis in applications such as telepresence and interactive agents. However, most existing methods focus on person-centric motion-driven generation, without explicitly modeling human-object interaction or its effects on the surrounding local scene. In everyday interaction, a person not only exhibits coordinated body, hand, and facial dynamics, but also alters the nearby scene through contact and object manipulation. \cite{DBLP:conf/cvpr/BhatnagarX0STP22,DBLP:conf/eccv/XieBP22,DBLP:conf/eccv/LiCMWPL24,gao2026lomelearninghumanobjectmanipulation,wang2026hand2worldautoregressiveegocentricinteraction}. This broader perspective motivates \emph{human-centric world modeling}, where the goal is to synthesize coherent, controllable, and interactive local world states centered on a person.

Existing approaches address only parts of this problem. Motion-driven human generation methods, including human reenactment and avatar animation, primarily focus on transferring or reproducing human motion through explicit skeletal poses, keypoints, or implicit motion latents ~\cite{DBLP:conf/nips/SiarohinLT0S19,DBLP:journals/corr/abs-2407-03168,DBLP:conf/iclr/ZhaoXSXZ0LSL25,DBLP:journals/corr/abs-2602-07498,DBLP:journals/corr/abs-2602-03796,cheng2025wananimateunifiedcharacteranimation,DBLP:conf/siggrapha/SongXZXGLZL25}. While effective for human motion synthesis, they typically model the person in isolation and therefore do not explicitly capture human-object contact or the resulting changes in the surrounding local scene. In addition, their control signals often remain insufficient for fine-grained manipulation-critical dynamics, particularly in the hands and face~\cite{PIXIE2021,DBLP:conf/aaai/MaHCWC0C24,DBLP:journals/corr/abs-2502-06145}. In parallel, recent text-conditioned image and video generation models ~\cite{DBLP:conf/siggrapha/ChristenHSRHSMT24,DBLP:journals/corr/abs-2411-17383,wan2025,DBLP:conf/iclr/YangTZ00XYHZFYZ25,kong2025hunyuanvideosystematicframeworklarge,DBLP:journals/corr/abs-2503-07598,DBLP:journals/corr/abs-2509-08519,DBLP:journals/corr/abs-2602-01538,DBLP:journals/corr/abs-2601-10103,DBLP:journals/corr/abs-2506-10568} have shown strong semantic generation ability, including prompt-specified human-object interactions. However, because these methods rely on free-form prompts, human motion and interaction semantics are often entangled in the same text signal, leading to weak motion control and unstable or inconsistent contact outcomes. Finally, both classes of methods remain computationally expensive and are largely limited to offline generation, making them unsuitable for real-time interactive deployment. As a result, existing techniques still lack a unified framework that jointly supports fine-grained human-state control, explicit object interaction-state control, and real-time performance.

In this work, we present a human-centric world model for upper-body interactive generation. Rather than targeting a general physical world \cite{robbyantteam2026advancingopensourceworldmodels,DBLP:conf/icml/BruceDEPS0LMSAA24}, we focus on a controllable local visual world centered on an upper-body human and the nearby objects involved in interaction. In this setting, the generated scene is determined by two complementary components: the \emph{human state}, which captures coordinated upper-body, hand, and facial dynamics as a continuous control variable, and the \emph{discrete interaction state}, which specifies contact-related relations between the person and surrounding objects as a discrete control variable. This perspective treats interaction as a central aspect of generation: the person is not only animated, but can also interact with nearby objects and affect the local environment. By jointly modeling human dynamics and object interaction states, our framework supports state-conditioned synthesis of person-centered scene evolution and is well suited to real-time interactive applications.

To support this formulation, our framework combines two corresponding forms of control. For human-state modeling, we introduce a unified implicit representation based on multi-scale motion encoding \cite{DBLP:conf/siggrapha/SongXZXGLZL25,DBLP:journals/corr/abs-2602-03796}, which extracts and fuses motion latents from upper-body, hand, and facial regions into a single latent space. This design enables expressive and fine-grained control of coordinated body-hand-face dynamics without explicit retargeting while remaining compact and reducing identity leakage. For interaction-state modeling, we represent interaction using language-conditioned \emph{interaction-state commands}, rather than free-form semantic prompts. This design is motivated by the structure of human-environment interaction: under similar human motion, a person may either contact an object or remain non-contact, making interaction naturally representable as a discrete state. Since human motion is already specified by the implicit human-state latent, it is sufficient to define discrete contact-oriented interaction states and combine them with the continuous control signal to characterize both motion and interaction. We therefore use language only to specify scene-agnostic discrete interaction states, such as \emph{no contact} and \emph{grasp}, as a complement to continuous human-state control. To facilitate learning under this formulation, we further build a dedicated rendering pipeline for human-object interaction data with aligned discrete state supervision. Together, these components form a continuous-discrete joint control scheme that models both human motion and interaction-driven local scene changes.

To enable real-time interaction, we accelerate the model with CausVid \cite{yin2024improveddistributionmatchingdistillation,yin2025slowbidirectionalfastautoregressive} and Self-Forcing \cite{huang2025selfforcingbridgingtraintest}, achieving streaming real-time generation at 25 FPS on two H100 GPUs. Experiments show that our method generates more coherent and realistic human-environment interactions than prior baselines, particularly in hand-object coordination and contact-aware local scene changes, 
Overall, our work takes a practical step beyond motion imitation toward real-time interactive modeling of human-centered local world states. Our contributions are three-fold:
\begin{itemize}
\item We present a human-centric world model for upper-body interactive generation, which formulates person-centered scene evolution through the joint modeling of continuous human states and discrete interaction states.
\item We introduce a unified implicit multi-scale representation for human-state control, together with language-conditioned discrete interaction-state commands, enabling controllable human-object interaction generation.
\item We develop the framework into a real-time interactive system, enabling streaming generation at 25 FPS on H100 GPUs.
\end{itemize}

\section{Related Work}

\subsection{Human Interaction Generation and Control}
Realistic human interaction requires modeling both human motion and human-object contact, but prior work has largely treated them separately. Motion-driven human video generation methods typically animate a reference person. \cite{DBLP:conf/iccv/KwonTSBP21,DBLP:conf/bmvc/MittalZT11,DBLP:conf/cvpr/ShanGSF20,DBLP:conf/cvpr/ZhangDGCABHT25,DBLP:conf/cvpr/CoronaPAMR20,DBLP:conf/3dim/KarunratanakulY20,DBLP:conf/eccv/LuKLLYHH24,DBLP:conf/cvpr/PrakashLAF0S25,DBLP:conf/cvpr/YeL0MBSTL23,DBLP:conf/cvpr/ZhangFDLT024,DBLP:conf/eccv/LiCMWPL24,DBLP:conf/cvpr/ZhongJYM25,DBLP:journals/corr/abs-2506-21552,fan2025reholdvideohandobject, DBLP:journals/corr/abs-2411-17383,DBLP:journals/corr/abs-2506-09984,DBLP:conf/nips/SiarohinLT0S19}. While effective for human animation, these methods primarily focus on the person itself and usually do not explicitly model interaction outcomes in the local scene, especially in upper-body settings where hand articulation and facial dynamics are important for realistic contact-aware generation. \cite{DBLP:journals/corr/abs-2407-03168,DBLP:conf/iclr/ZhaoXSXZ0LSL25,DBLP:conf/siggrapha/MaLWPHYZCS0C24,DBLP:conf/siggrapha/SongXZXGLZL25,DBLP:journals/corr/abs-2502-06145,cheng2025wananimateunifiedcharacteranimation,klingteam2026klingmotioncontroltechnicalreport,DBLP:journals/corr/abs-2602-07498,DBLP:journals/corr/abs-2602-03796,PIXIE2021}.

On the other hand, recent image and video generation models have demonstrated the ability to depict human-object interactions through text conditioning \cite{DBLP:conf/siggrapha/ChristenHSRHSMT24,DBLP:conf/nips/WeiJXT00C024,DBLP:journals/corr/abs-2601-10103,DBLP:journals/corr/abs-2503-07598,wan2025,hacohen2026ltx2efficientjointaudiovisual,kong2025hunyuanvideosystematicframeworklarge,DBLP:journals/corr/abs-2602-01538,DBLP:conf/iclr/YangTZ00XYHZFYZ25,DBLP:conf/aaai/MouWXW0QS24}. However, these approaches rely on free-form prompts, making interaction states difficult to control explicitly over time \cite{DBLP:journals/corr/abs-2503-07598,DBLP:journals/corr/abs-2602-01538,DBLP:journals/corr/abs-2509-08519}. This is especially limiting in interactive settings, where low-latency state switching is required. Human--object interaction, grasping, contact modeling \cite{DBLP:conf/cvpr/TamuraOY21,DBLP:conf/cvpr/KimLKKK21,chao2018learningdetecthumanobjectinteractions,DBLP:conf/cvpr/BhatnagarX0STP22,DBLP:conf/eccv/XieBP22,DBLP:conf/eccv/QinWLJYFW22,DBLP:conf/cvpr/ChaoYXMHTNWIBKF21,DBLP:conf/iccv/MoGM0T21} study interaction more directly, but are mainly designed for recognition, prediction, geometry, or robotics rather than controllable person-centered video generation. They also rarely treat interaction as an explicit control variable for generation. 

In contrast, we model interaction with a continuous-discrete control scheme: a unified human-state representation for motion and discrete interaction states for contact. This separation enables more direct and controllable human-centric interaction generation.

\subsection{Interactive World Modeling}
Recent world models and generative simulators have shown strong ability to model long-horizon scene evolution and generate open-ended environments.
\cite{DBLP:conf/icml/BruceDEPS0LMSAA24,DBLP:journals/corr/abs-2508-13009,DBLP:journals/corr/abs-2506-17201,DBLP:journals/corr/abs-2506-09995,DBLP:conf/cvpr/BarZTDL25,robbyantteam2026advancingopensourceworldmodels,DBLP:journals/corr/abs-2602-22960,huang2026pointworldscaling3dworld,DBLP:journals/corr/abs-2404-02101,DBLP:journals/corr/abs-2309-17080}. These methods aim to support interactive exploration of generated scenes over time, moving beyond static image or video synthesis toward persistent visual worlds. However, current world models remain largely exploration-oriented. In most cases, interaction is limited to viewpoint control, such as navigating a virtual camera, rather than explicitly editing or altering local scene states \cite{robbyantteam2026advancingopensourceworldmodels}. They also typically do not support fine-grained human motion control, making them less suitable for scenarios where detailed human actions should drive local scene changes through interaction.

In parallel, extensive work has improved the efficiency of diffusion and video generation through distillation, consistency learning, reduced-step sampling, and streaming generation \cite{song2023consistencymodels,yin2025slowbidirectionalfastautoregressive,yin2024improveddistributionmatchingdistillation,huang2025selfforcingbridgingtraintest,zhu2026causalforcingautoregressivediffusion}. While these methods support real-time deployment, they are not designed for interaction-aware generation with explicit control over human motion and object interaction states. Our work instead targets a controllable local visual world centered on an upper-body human, and leverages these efficiency techniques to enable real-time generation under online state switching.

\section{Method}
\subsection{Overview and Problem Formulation}

\begin{figure*}[htb]
    \centering
    \includegraphics[width=\textwidth]{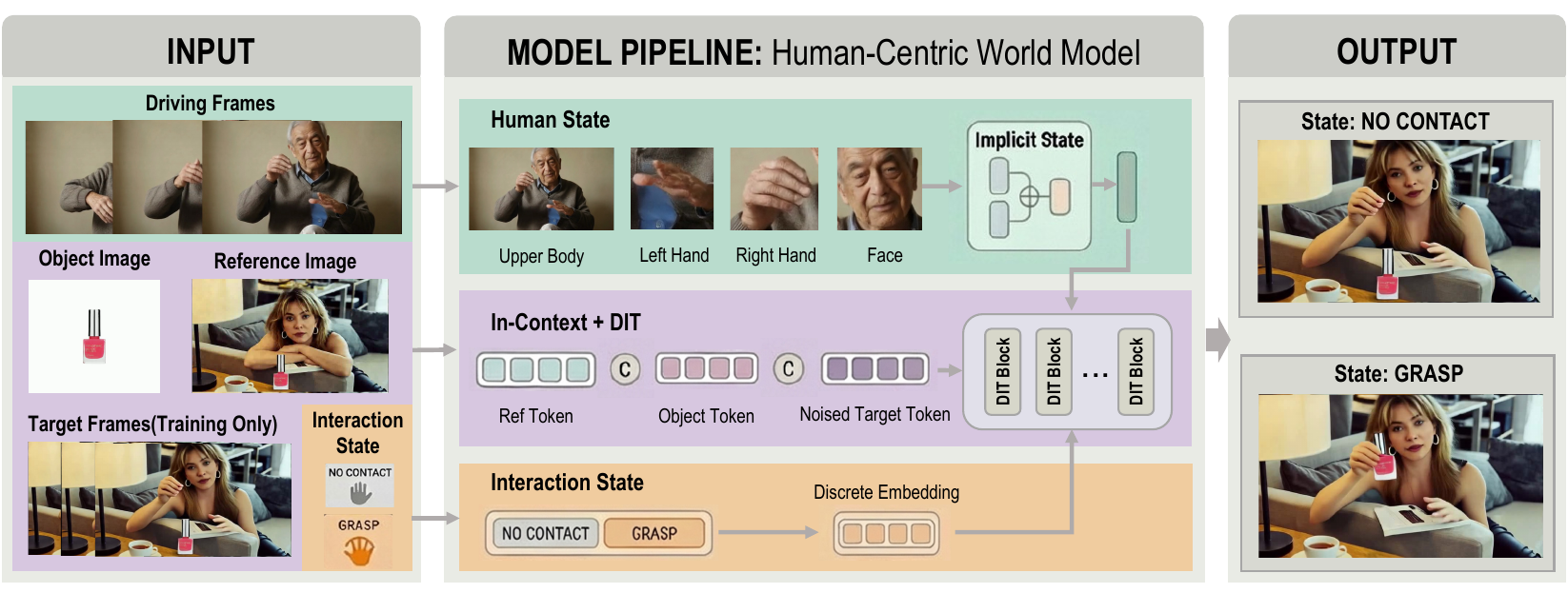}
    \caption{\textbf{An overview of our framework}. Given a reference image, an object reference image, a driving signal, and a discrete interaction-state command, the Human State module captures the person's multi-scale implicit motion states, while the Interaction State module models the human–object contact states. These two states are then jointly fed into DiT to synthesize the final result.
    }
    \label{fig:pipeline}
\end{figure*}

We formulate upper-body interactive generation as a \emph{human-centric world modeling} problem. Rather than modeling a general physical world, we focus on a controllable local visual world centered on an upper-body human and the nearby objects involved in interaction. In this setting, generation depends not only on how the person moves, but also on how the person interacts with the environment. The overall framework is shown in Fig.\ref{fig:pipeline}.

We represent this process using two complementary state variables: a \emph{human state} $\mathbf{s}^{h}$ and an \emph{discrete interaction state} $\mathbf{s}^{o}$. The human state $\mathbf{s}^{h}$ captures coordinated upper-body, hand, and facial dynamics, and is represented as a unified implicit latent. The interaction state $\mathbf{s}^{o}$ specifies the contact-related relation between the person and nearby objects. We model $\mathbf{s}^{o}$ as discrete interaction-state command: \textit{no contact} and \textit{grasp}. The two states play complementary roles: $\mathbf{s}^{h}$ determines how the person moves, while $\mathbf{s}^{o}$ determines how the person engages with surrounding objects. 

Formally, given a reference appearance input $\mathbf{x}_{\mathrm{ref}}$, an object image $\mathbf{x}_{\mathrm{obj}}$, a driving motion input $\mathrm{d}$, and a discrete interaction-state command $\mathbf{s}^{o}$ encoded by a text encoder, our model first derives the human-state latent

\begin{equation}
\begin{aligned}
    \mathbf{s}^{h} = E(\mathrm{d})
\end{aligned}
\end{equation}

where $E(\cdot)$ denotes the human-state encoder. The output frame or video $\hat{\mathbf{y}}$ is then synthesized as
\begin{equation}
\label{eq:composite}
\begin{aligned}
    \hat{\mathbf{y}} = G(\mathbf{x}_{\mathrm{ref}}, \mathbf{x}_{\mathrm{obj}}, \mathbf{s}^{h}, \mathbf{s}^{o}),
\end{aligned}
\end{equation}
where $G(\cdot)$ is the generator. Under this formulation, generation is jointly governed by continuous human-state control and discrete interaction-state control, allowing the model to synthesize local scene evolution as a person moves and interacts with nearby objects.

\subsection{Human State Modeling}

Human motion is central to upper-body interactive generation. We represent it with an implicit multi-scale latent, $\mathbf{s}^{h}$, that captures upper-body dynamics in a unified manner. Specifically, $\mathbf{s}^{h}$ models both global upper-body motion and fine-grained local dynamics of the hands and face, yielding a coordinated representation of upper-body, hand, and facial motion. Compared with explicit controls such as skeletons or keypoints, this implicit representation is more expressive, preserves subtle motion details, and yields better cross-identity motion transfer.

This design is motivated by the multi-scale nature of upper-body motion. The upper body mainly reflects coarse pose and torso dynamics, whereas the hands and face contain much finer motion patterns, such as finger articulation and facial expression changes. A single global latent often under-represents such local details. We therefore adopt a multi-scale motion encoding scheme to enhance local fidelity while preserving global coordination.

\begin{figure}[t]
    \centering
    \includegraphics[width=\columnwidth]{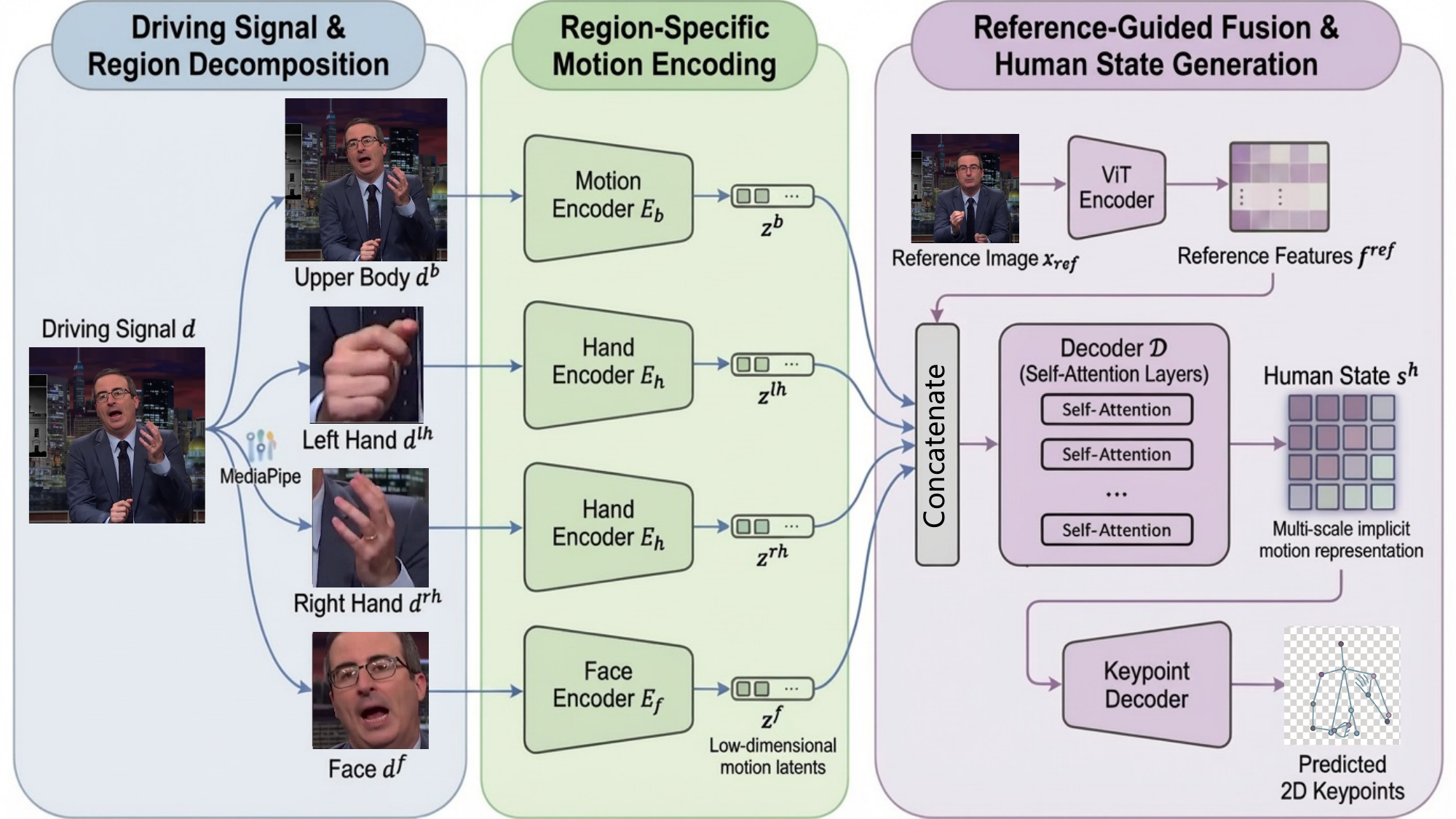}
    \caption{Overview of Human State generation, which decomposes the driving signal into body-part regions, encodes region-specific motions, and fuses them with reference features to produce a multi-scale implicit human state.}
    \label{fig:humanState}
\end{figure}

As shown in Fig.\ref{fig:humanState}, given a driving signal $\mathbf{d}$, we extract region-level motion inputs from four complementary regions: the upper body, the left hand, the right hand, and the face, denoted by $\mathbf{d}^{b},\mathbf{d}^{lh},\mathbf{d}^{rh}, \mathbf{d}^{f}$. To obtain the local regions, we use a MediaPipe \cite{lugaresi2019mediapipeframeworkbuildingperception} detector to estimate the bounding boxes of the hands and face, and crop the corresponding image regions from the driving input. Here, $\mathbf{d}^{b}$ mainly captures global upper-body dynamics, while $\mathbf{d}^{lh}$, $\mathbf{d}^{rh}$, and $\mathbf{d}^{f}$ represent the left hand, right hand, and facial regions, respectively. These inputs are then encoded into latent motion features:
\begin{equation}
\begin{aligned}
    \mathbf{z}^{b} = E_b(\mathbf{d}^{b}) \quad
    \mathbf{z}^{lh} = E_h(\mathbf{d}^{lh}) \quad
    \mathbf{z}^{rh} = E_h(\mathbf{d}^{rh}) \quad
    \mathbf{z}^{f} = E_f(\mathbf{d}^{f})
\end{aligned}
\end{equation}

where $E_b(\cdot)$, $E_h(\cdot)$, and $E_f(\cdot)$ denote the upper-body, hand, and face motion encoders, respectively. All three encoders share the same ViT-based architecture \cite{dosovitskiy2021imageworth16x16words} and map their input image regions to 1D latent vectors. Following the information bottleneck principle, we keep these vectors low-dimensional to reduce identity leakage and retain only motion-relevant information.

We then fuse the motion latents and implicitly retarget them to the reference identity following X-Unimotion \cite{DBLP:conf/siggrapha/SongXZXGLZL25}. Specifically, a ViT encoder extracts reference image patches from the reference image, denoted by $\mathbf{f}^{\mathrm{ref}} \in \mathbb{R}^{(h \times w) \times c}$. The motion latents are concatenated with linearly projected reference image patches and fed into a decoder $D$ \cite{he2021maskedautoencodersscalablevision} composed of multiple self-attention layers. The decoder outputs a spatially aligned feature map, which serves as the final human motion representation. This design integrates multi-scale motion cues while aligning them with the reference appearance. Finally, the human state is defined as
\begin{equation}
\begin{aligned}
    \mathbf{s}^{h} = D\big(\mathbf{f}^{\mathrm{ref}}, \mathbf{z}^{b}, \mathbf{z}^{lh}, \mathbf{z}^{rh}, \mathbf{z}^{f}\big).
\end{aligned}
\end{equation}
The resulting representation integrates fine-grained motion cues with implicit retargeting to the reference identity, making $\mathbf{s}^{h}$ a compact and effective control signal for generation. Through this implicit retargeting, $\mathbf{s}^{h}$ can be combined with a reference image to synthesize identity-consistent motion-driven outputs.

\subsection{Discrete Interaction-State Modeling}
\label{sec:obj}

Beyond human motion, human--object interaction is another key factor in upper-body interactive generation. Similar human motion can correspond to different contact relations with nearby objects, leading to different local visual outcomes. We therefore introduce a discrete interaction state, denoted by $\mathbf{s}^o$, to model the contact relation between the person and surrounding objects.

Human--object interaction naturally has a discrete state structure. Under similar human motion, a person may either grasp a nearby object or remain disengaged, and this difference is more naturally described as a change in interaction state than as a change in motion itself. Motivated by this observation, we represent interaction as a set of discrete interaction-states instead of open-ended semantic descriptions. Specifically, we define the object interaction state as $\mathbf{s}^o \in \mathcal{S}$, where $\mathcal{S}$ denotes the set of discrete interaction states. In practice, each interaction state is represented by a short fixed text phrase, such as ``no contact'' or ``grasp'', and encoded by a frozen T5 text encoder. In this work, we consider two basic states:
\begin{equation}
\begin{aligned}
    \mathcal{S} = \{\textit{no contact},\ \textit{grasp}\}
\end{aligned}
\end{equation}
Here, \textit{no contact} denotes the absence of contact with the target object, while \textit{grasp} denotes a grasping relation with the object.

Our formulation differs fundamentally from conventional interaction generation methods. First, language is not used as a task-specific free-form description, but as a unified abstract interaction-state command with only a small number of discrete variables, avoiding the need for scene-specific textual prompts. Second, since human motion is already specified by the human state $\mathbf{s}^h$, it is sufficient to define discrete interaction states and combine them with this continuous control signal to jointly characterize motion and interaction. In this way, interaction control remains explicit without introducing redundant semantic descriptions. Finally, this discrete state formulation is naturally suited to real-time interactive scenarios, as state switches can be parsed and applied immediately, enabling online interruption and transition during motion. During generation, the interaction state $\mathbf{s}^{o}$ and the human state $\mathbf{s}^{h}$ jointly determine the resulting local visual state. 

\begin{figure}[t]
    \centering
    \includegraphics[width=\columnwidth]{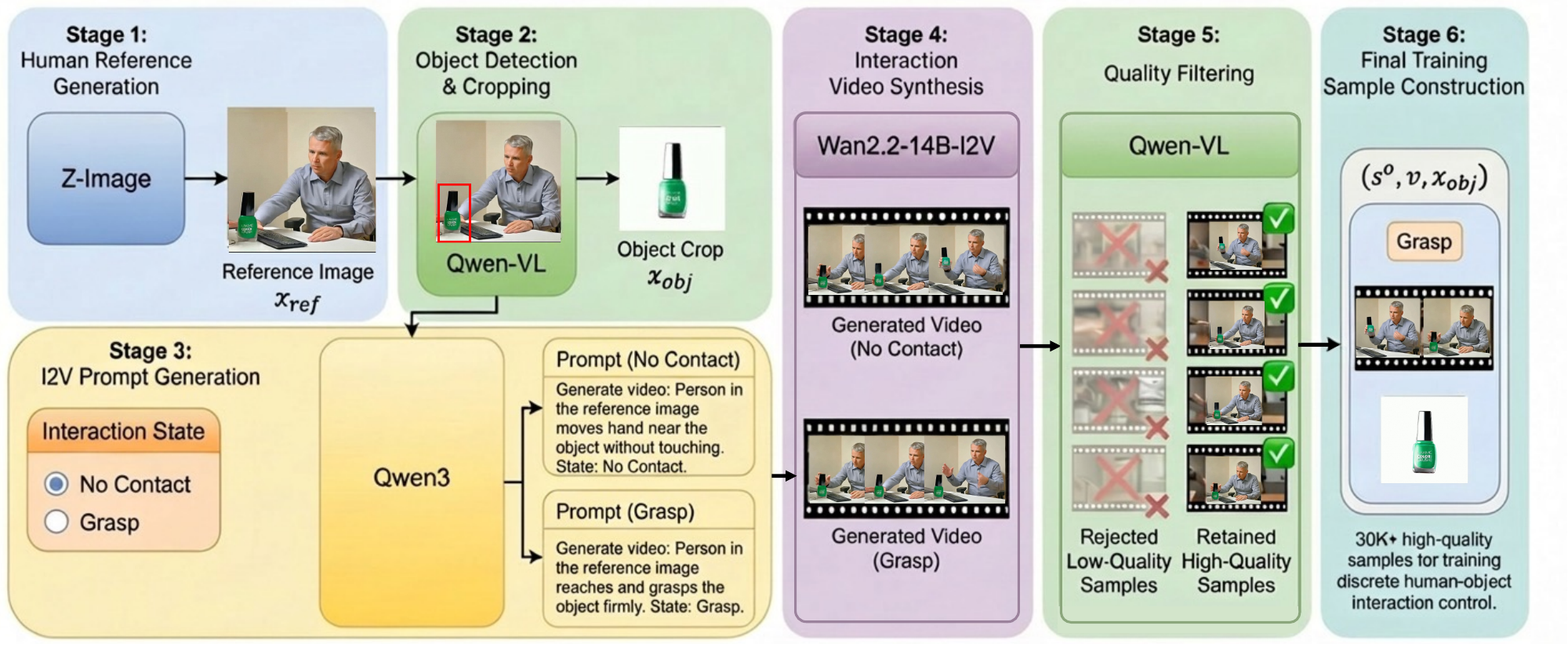}
    \caption{Pipeline for constructing a high-quality human–object interaction video dataset with discrete contact states (e.g., no contact and grasp).}
    \label{fig:dataGen}
\end{figure}

\noindent
\textbf{Data production pipeline}
To train the discrete interaction-state control, we construct a dedicated data production pipeline, as shown in Fig.\ref{fig:dataGen}. We first use Z-Image \cite{imageteam2025zimageefficientimagegeneration} to generate upper-body human images as reference images. For each reference image, we apply Qwen-VL \cite{bai2025qwen3vltechnicalreport} to detect tabletop objects and crop the corresponding object reference image, denoted by $\mathbf{x}_{obj}$. We then use Qwen3 \cite{yang2025qwen3technicalreport} to generate image-to-video prompts conditioned on the reference image and the target object. To cover different interaction modes, we construct two prompt types corresponding to the \textit{grasp} and \textit{no contact} states. Given the prompt and the reference image, Wan2.2-14B-I2V \cite{wan2025} is used to synthesize a human-object interaction video $\mathbf{v}$.

To improve data quality, we further apply Qwen-VL to filter out low-quality results and retain only high-quality samples. We also manually inspect a subset of the collected samples to verify interaction plausibility and visual quality. Using this pipeline, we collect over 30K samples. Each sample consists of an interaction-state label, a video, and an object reference image, i.e.,
$(\mathbf{s}^{o}, \mathbf{v}, \mathbf{x}_{obj})$.
We will release this dataset to support future research on controllable human-object interaction generation.

\begin{figure*}[htb]
    \centering
    \includegraphics[width=\textwidth]{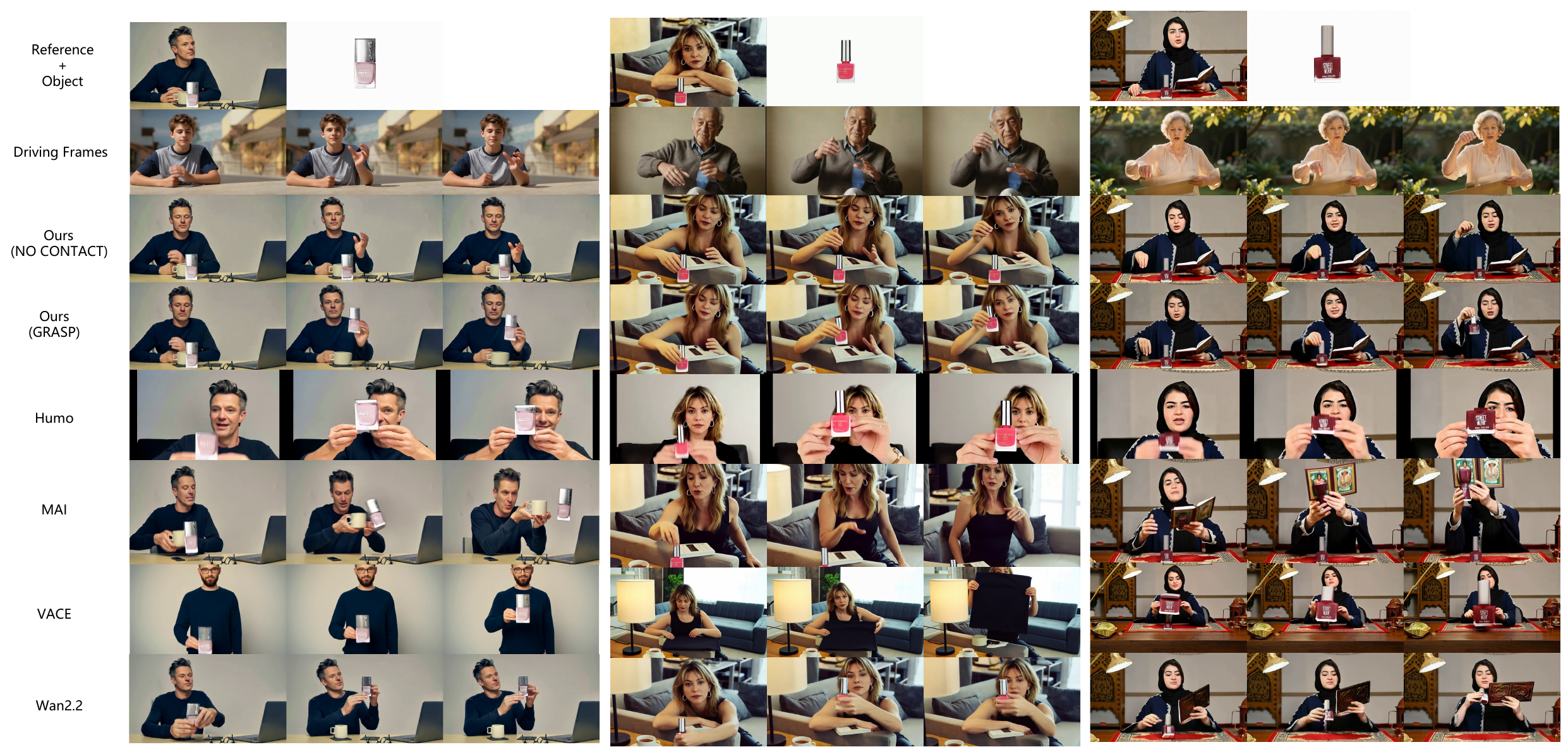}
    \caption{Qualitative comparison with state-of-the-art methods. Please refer to the supplementary material for more results.
    }
    \label{fig:qualitative_comparison}
\end{figure*}

\begin{table*}[t]
\centering
\caption{\textbf{Quantitative comparison on objective and subjective metrics.}}
\label{tab:main_results}
\small
\setlength{\tabcolsep}{5pt}
\renewcommand{\arraystretch}{1.15}
\begin{tabular}{lccccccccccc}
\toprule
& \multicolumn{4}{c}{Global Metrics} & \multicolumn{3}{c}{Local Similarity} & \multicolumn{4}{c}{User Study} \\
\cmidrule(lr){2-5} \cmidrule(lr){6-8} \cmidrule(lr){9-12}
Method 
& FVD $\downarrow$ 
& tLPIPS $\downarrow$ 
& CLIP-I $\uparrow$ 
& OQ $\uparrow$
& HandSIM $\downarrow$ 
& FaceSIM $\downarrow$ 
& IdentitySIM $\uparrow$
& TF $\uparrow$ 
& AF $\uparrow$ 
& MC $\uparrow$ 
& VQ $\uparrow$ \\
\midrule
Wan-Animate    & 182.4 & 0.257 & 0.836 & 0.108 & 0.060 & 0.051 & 0.801 & 3.28 & \underline{3.56} & \underline{3.47} & 2.74 \\
Wan2.2-TI2V-5B & 156.8 & 0.178 & \underline{0.848} & 0.132 & -- & -- & 0.811 & 3.54 & 3.33 & 3.41 & 3.68 \\
Humo           & 178.9 & 0.193 & 0.829 & 0.103 & -- & -- & 0.776 & 2.91 & 3.05 & 2.98 & 3.12 \\
MAI            & \textbf{152.7} & \textbf{0.176} & 0.842 & \underline{0.147} & -- & -- & \underline{0.812} & \underline{3.76} & 3.41 & 3.35 & \textbf{3.91} \\
VACE           & 160.3 & 0.184 & 0.839 & 0.118 & -- & -- & 0.795 & 3.17 & 3.39 & 3.28 & 3.31 \\
Ours           & \underline{154.1} & \underline{0.178} & \textbf{0.851} & \textbf{0.151} & \textbf{0.049} & \textbf{0.041} & \textbf{0.829} & \textbf{4.08} & \textbf{3.94} & \textbf{3.88} & \underline{3.82} \\
\bottomrule
\end{tabular}
\end{table*}

\subsection{Real-Time Human-Centric World Synthesis}

Human--object interaction places strict demands on generation efficiency, especially in motion-driven interactive scenarios. To achieve streaming real-time inference, we further optimize the generation model for low-latency online synthesis. First, we distill the model with CausVid \cite{yin2025slowbidirectionalfastautoregressive}, reducing the sampling steps from 20 to 4 while introducing causal attention and sliding-window attention for efficient autoregressive video generation. This substantially improves inference speed without sacrificing temporal coherence. To further improve long-horizon stability, we adopt the Self-Forcing \cite{huang2025selfforcingbridgingtraintest} to mitigate temporal error accumulation during iterative generation. With these optimizations, our system achieves streaming real-time performance at 25 FPS on two H100 GPUs for 512$\times$512 generation, with an end-to-end latency of about 1000 ms.

\subsection{Two-Stage Training Strategy}

We adopt a two-stage training strategy to first learn a stable human-state representation and then incorporate interaction-state control.

\noindent
\textbf{Stage I: Human-state pretraining.}
In the first stage, we learn the unified implicit human representation from general upper-body human videos without object reference images or interaction-state supervision. The text condition is set to null so that the model focuses solely on human motion dynamics. In practice, training with only the Flow Matching objective $\mathcal{L}_{\mathrm{FM}}$ converges slowly. We therefore introduce two auxiliary losses: a representation alignment loss $\mathcal{L}_{\mathrm{RePA}}$ that aligns intermediate DiT features with DINOv3~\cite{simeoni2025dinov3} features to encourage structured visual representations, and a keypoint supervision loss $\mathcal{L}_{\mathrm{kp}}$ through a lightweight decoder $D_k$, which predicts target 2D keypoints from the implicit motion variable under ground-truth supervision to provide geometric guidance. The training objective is:

\[
\mathcal{L}_{\text{stage1}}
=
\mathcal{L}_{\mathrm{FM}}
+
\lambda_{\mathrm{rep}} \mathcal{L}_{\mathrm{RePA}}
+
\lambda_{\mathrm{kp}} \mathcal{L}_{\mathrm{kp}},
\]
where $\lambda_{\mathrm{rep}}$ is set to 0.1 and $\lambda_{\mathrm{kp}}$ is set to 10.

\noindent
\textbf{Stage II: Interaction-state learning.}
In the second stage, we initialize the model from Stage I and introduce discrete interaction-state commands together with object reference images. Training is performed on the rendered interaction dataset. To preserve the learned human-state representation, we freeze the pretrained motion encoder and optimize the remaining modules for interaction-aware generation. We use only the Flow Matching objective for Stage II:
\[
\mathcal{L}_{\text{stage2}} = \mathcal{L}_{\mathrm{FM}}.
\]

\section{Experiments}

\subsection{Implementation Details}

Our model is built upon Wan2.2-5B \cite{wan2025} and initialized from its pretrained weights. The motion encoder adopts a ViT-Base architecture, and the text encoder is based on T5. The training data consists of two parts. For human-state learning, we use monocular upper-body motion videos, including filtered clips from SpeakerVid-5M \cite{zhang2025speakervid5mlargescalehighqualitydataset} and an internal sign-language dataset, resulting in over 3 million training samples. For interaction-state learning, we use the synthetic dataset described in Sec.\ref{sec:obj}, which contains over 30K samples. During training, videos are processed at their original aspect ratios, and each sample contains 45 frames. Training proceeds in two stages. Stage I uses the human-state dataset for 100K steps on 128 A100 GPUs. Stage II initializes from the Stage I checkpoint, freezes the motion encoder, and trains on the interaction dataset for 10K steps on 32 A100 GPUs.

\noindent
\textbf{Baselines and Metrics.}
We compare our method with Wan2.2-5B \cite{wan2025}, Humo \cite{DBLP:journals/corr/abs-2509-08519}, VACE \cite{DBLP:journals/corr/abs-2503-07598} and MAI \cite{DBLP:journals/corr/abs-2602-01538} for interaction-aware generation. These baselines primarily rely on text-based conditioning and do not expose an explicit human-state control variable comparable to ours. To further evaluate the controllability of our human-state representation, we also compare with Wan-Animate \cite{cheng2025wananimateunifiedcharacteranimation}, a state-of-the-art open-source human animation method. For quantitative evaluation, we use FVD and tLPIPS to measure overall video quality and temporal coherence, while CLIP-I \cite{ruiz2023dreamboothfinetuningtexttoimage} and Object Quality (OQ) are used to evaluate appearance fidelity and object fidelity, respectively. Object Quality (OQ) is assessed by the product of object dynamics score~\cite{zhang2025speakervid5mlargescalehighqualitydataset} and object DINO consistency~\cite{DBLP:conf/iclr/0097LL000NS23}.
To assess fine-grained local quality, we report HandSIM and FaceSIM for hand and facial motion similarity, as well as IdentitySIM for identity preservation. HandSIM and FaceSIM are computed as the average per-frame MSE between the hand coefficients and facial blendshape coefficients extracted from generated and ground-truth videos using off-the-shelf hand~\cite{pavlakos2024reconstructing} and face~\cite{lugaresi2019mediapipeframeworkbuildingperception} trackers, respectively. IdentitySIM is computed using ArcFace~\cite{Deng_2022}. In addition, we conduct a user study on text-following (TF), action-following (AF), motion consistency (MC), and visual quality (VQ) following LOME \cite{gao2026lomelearninghumanobjectmanipulation}.

\subsection{Human--Object Interaction}
\noindent
\textbf{Qualitative Results.}
Fig.~\ref{fig:qualitative_comparison} compares our method with existing interaction-aware generation approaches. VACE and Humo show weaker identity preservation and less consistent scene context, while MAI sometimes produces unstable contact, such as floating objects or implausible grasps. Wan2.2 also struggles to preserve object identity consistently. In addition, these baselines mainly rely on text-based conditioning, making precise control over motion trajectories more difficult. By contrast, our method achieves more faithful identity preservation, more stable hand--object interaction, and more accurate motion following. The human-state representation enables precise upper-body motion control, while the discrete interaction-state commands allow explicit switching between different contact relations under similar motion conditions.

\noindent
\textbf{Quantitative Results} 
For interaction-aware evaluation, we use a held-out test set of 100 high-quality clips constructed with the data pipeline in Sec.~3.3, without overlap with the Stage-II training set in identities, objects, or prompts. For motion reenactment, we use 100 upper-body driving videos from SpeakerVid-5M and 100 randomly sampled reference images of different identities. We further report HandSIM and FaceSIM to evaluate fine-grained hand and facial motion accuracy. As reported in Tab.~\ref{tab:main_results}, our method shows favorable performance in maintaining reference identity, controlling human--object interaction, and modeling fine-grained local motion, particularly in the hand and facial regions.

\begin{figure}[t]
    \centering
    \includegraphics[width=\columnwidth]{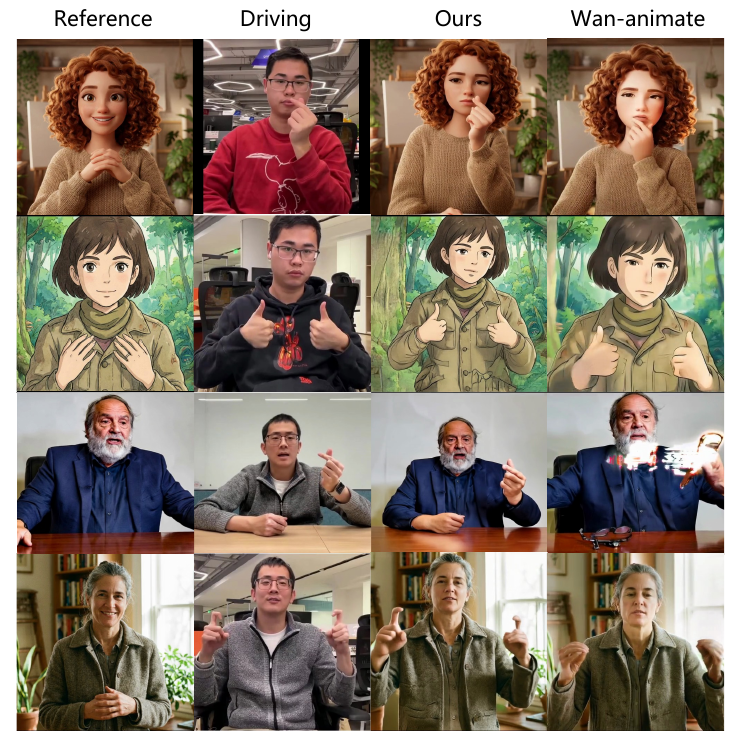}
    \caption{\textbf{Human State control.} Compared with Wan-Animate, our method achieves more accurate cross-identity driving and better local motion control, especially for complex hand movements.}
    \label{fig:human_state_control}
\end{figure}

\subsection{Human State}
Fig.~\ref{fig:human_state_control} demonstrates the effectiveness of Human State control. Compared with the open-source state-of-the-art baseline Wan-Animate, our method shows stronger cross-identity driving capability and more accurate control over local regions. In particular, for complex hand motions, our method more faithfully reproduces the target hand dynamics, whereas Wan-Animate is more affected by errors introduced during skeleton retargeting. The quantitative results in Tab.~\ref{tab:main_results}
further support this observation, showing that our method achieves lower hand-driving error.

\begin{figure}[t]
    \centering
    \includegraphics[width=\columnwidth]{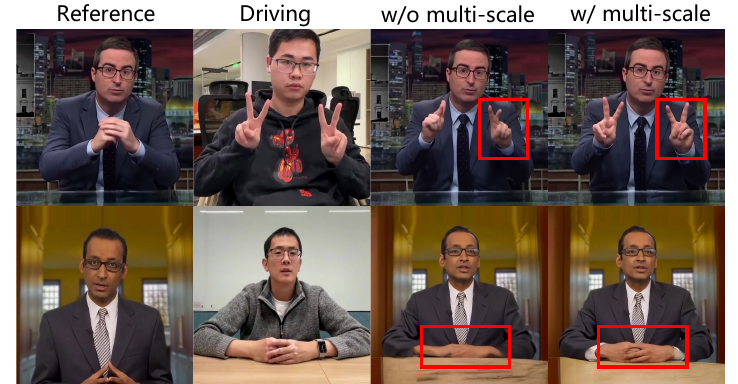}
    \caption{\textbf{Ablation study of the multi-scale Human State representation.}. \textit{w/o multi-scale} does not inject local hand and facial regions, and \textit{w/ multi-scale} is Human State modeling. Red boxes mark local regions.}
    \label{fig:abalation_human}
\end{figure}

\begin{figure}[t]
    \centering
    \includegraphics[width=\columnwidth]{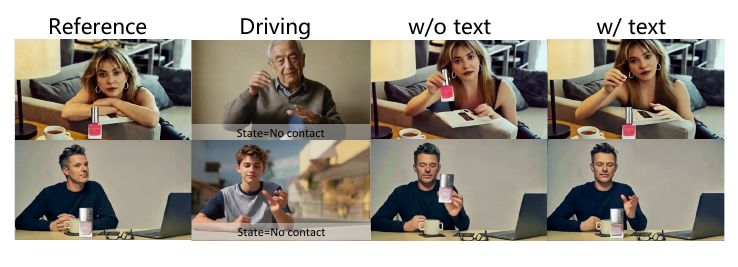}
    \caption{\textbf{Ablation study of interaction-state conditioning.} Comparison between \textit{w/o text}, which removes discrete text state control, and \textit{w/ text} interaction-state conditioning.}
    \label{fig:abalation_text}
\end{figure}

\subsection{Ablation Study}

\noindent
\textbf{Multi-scale human state.}
We compare a single-scale human state with our multi-scale design. As shown in Fig.\ref{fig:abalation_human}, the single-scale variant tends to provide weaker control over fine-grained regions, particularly the face and hands. In contrast, our design incorporates motion cues from multiple scales and applies dedicated encoders to local regions, which improves the controllability of local details.

\noindent
\textbf{Discrete Interaction State} We compare two interaction conditioning schemes: no interaction text and our discrete interaction-state commands, as shown in Fig.~\ref{fig:abalation_text}. Without explicit interaction conditioning, the model tends to generate grasping behavior by default, suggesting that object contact is heavily biased by the dataset distribution and cannot be controlled in a user-specified manner. In contrast, our discrete interaction-state commands provide explicit control over the interaction state, enabling more predictable generation of no-contact and grasp behaviors.

\section{Conclusion}
We presented a real-time human-centric local world model for upper-body human-object interaction generation. Our framework formulates person-centered scene evolution through the joint control of a continuous human state and a discrete interaction state, enabling coordinated modeling of upper-body, hand, and facial motion together with controllable human--object interaction. By introducing a unified multi-scale implicit human-state representation and discrete interaction-state commands, the proposed method improves identity-consistent motion control and contact-aware generation in local human-centered scenes. We further developed the framework into a streaming system with real-time inference, supporting online interaction-state switching at 25 FPS. Experimental results show favorable identity consistency, human--object interaction control, and fine-grained local motion modeling, advancing real-time human-centric world modeling.

\bibliographystyle{ACM-Reference-Format}
\bibliography{sample-base}

\clearpage
\appendix

\begin{figure*}[htb]
    \centering
    \includegraphics[width=\textwidth]{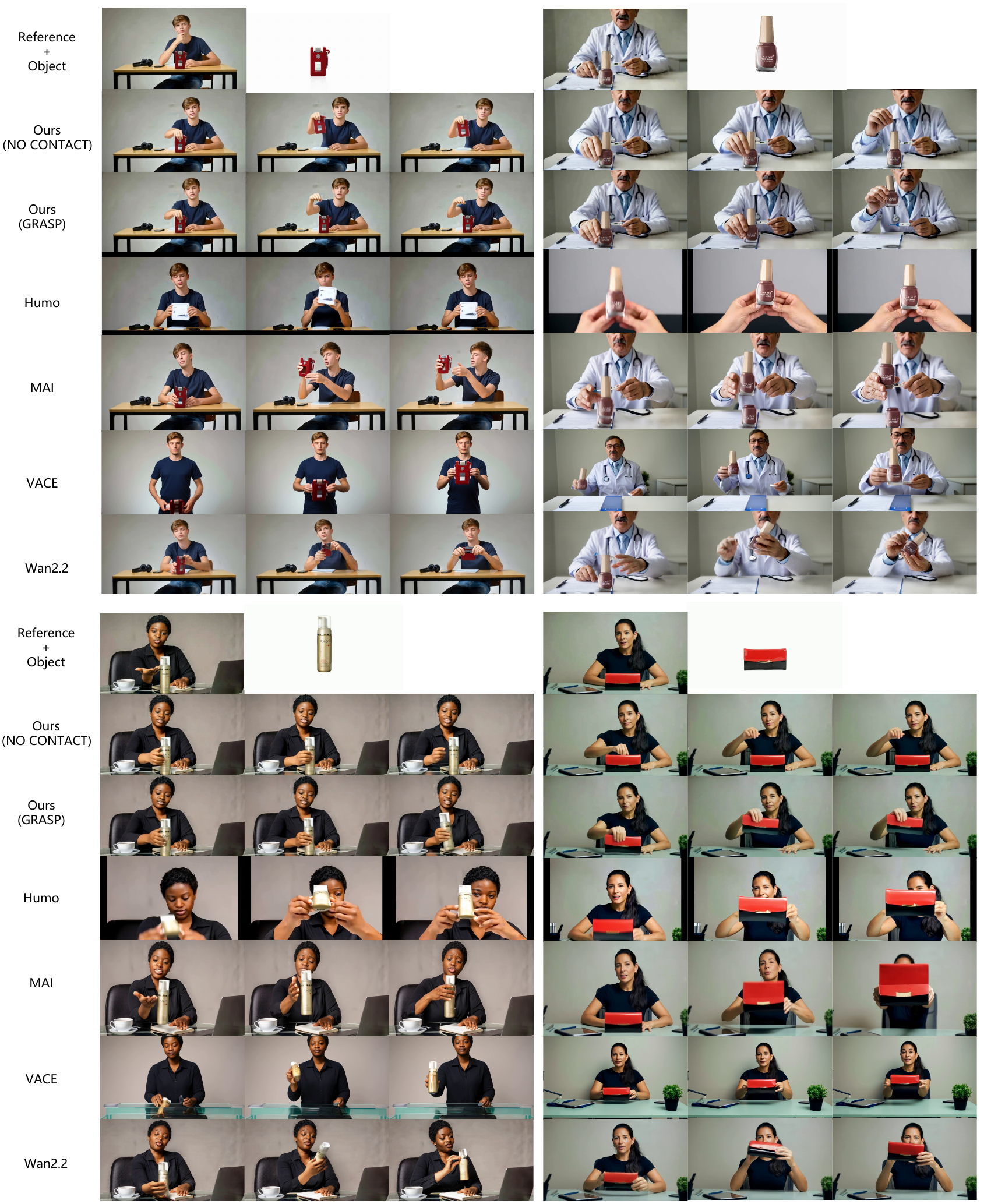}
    \caption{Extra visual results.
    }
    \label{fig:appendix}
\end{figure*}

\begin{figure*}[htb]
    \centering
    \includegraphics[width=\textwidth]{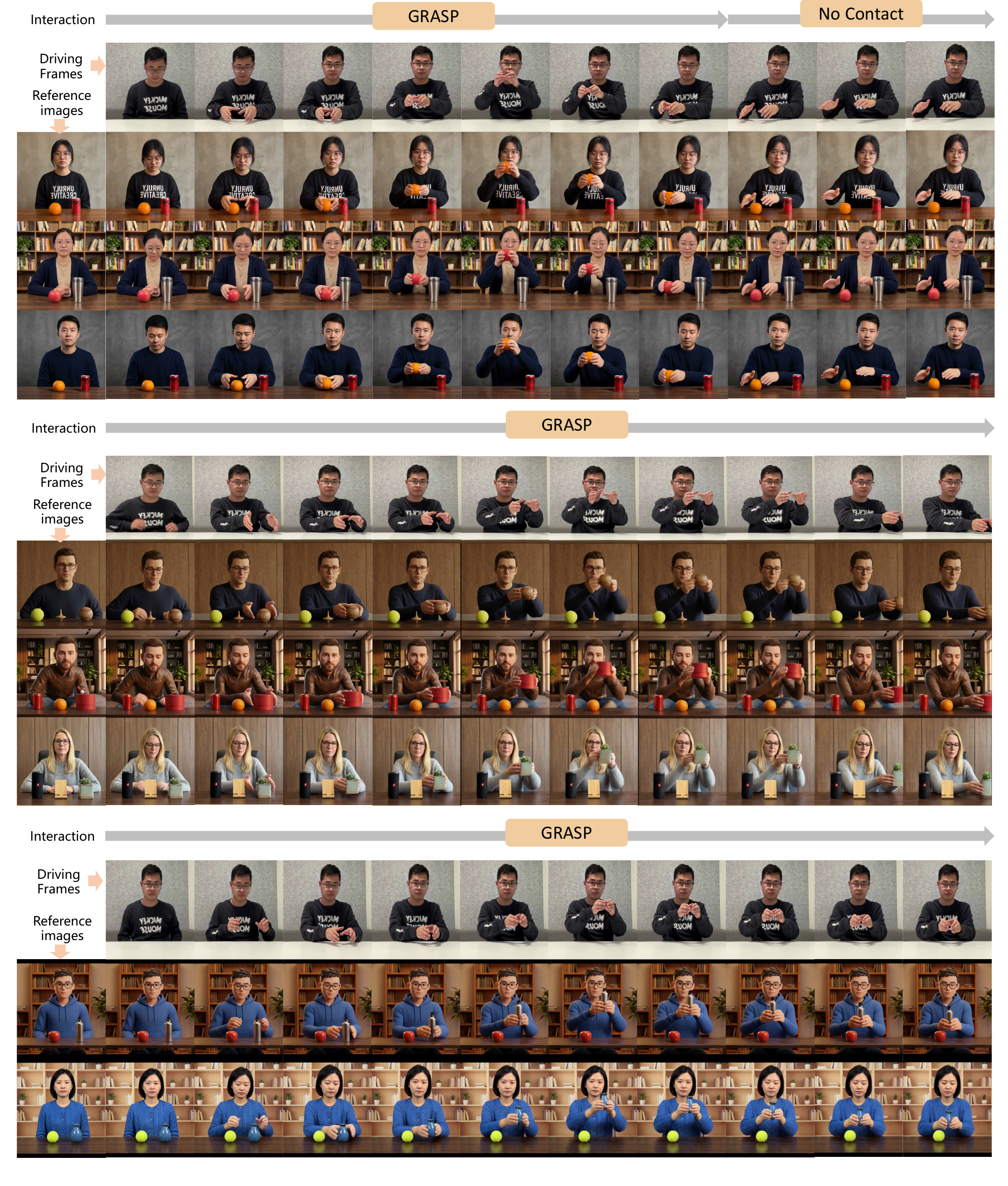}
    \caption{Additional results of streaming real-time driving. Given live driving input, our method generates temporally coherent upper-body interaction videos with stable identity preservation and controllable interaction states. The reference images are generated by Gemini.}
    \label{fig:appendix_streaming}
\end{figure*}

\end{document}